\documentclass[10pt]{article}

\usepackage[margin=0.9in,top=0.8in,bottom=0.8in]{geometry}
\usepackage{iftex}
\ifPDFTeX
  \usepackage[utf8]{inputenc}
  \usepackage[T1]{fontenc}
  \usepackage{lmodern}
\else
  \usepackage{fontspec}
\fi
\ifPDFTeX
  \newcommand{\authoremail}[1]{{\small\fontfamily{lmtt}\fontseries{l}\selectfont #1}}
\else
  \newfontfamily\emaillight{lmmonolt10-regular.otf}
  \newcommand{\authoremail}[1]{{\small\emaillight #1}}
\fi
\usepackage{microtype}
\usepackage{amsmath,amssymb,amsthm}
\usepackage{booktabs}
\usepackage{multirow}
\usepackage{graphicx}
\usepackage{xcolor}
\usepackage{enumitem}
\usepackage[font=small,labelfont=bf]{caption}
\usepackage[colorlinks=true,linkcolor=blue!60!black,citecolor=blue!60!black,urlcolor=blue!60!black]{hyperref}
\usepackage[numbers,sort&compress]{natbib}
\usepackage{listings}
\usepackage{tikz}
\usetikzlibrary{positioning,arrows.meta,shapes.geometric}
\usepackage{pifont}
\usepackage{needspace}
\usepackage{placeins}
\usepackage[compact]{titlesec}
\titlespacing*{\section}{0pt}{1.35ex plus .2ex minus .15ex}{0.75ex plus .15ex}
\titlespacing*{\subsection}{0pt}{1.15ex plus .2ex minus .15ex}{0.5ex plus .15ex}
\titlespacing*{\paragraph}{0pt}{0.9ex plus .15ex minus .1ex}{0.8em}
\ifPDFTeX\fi%
\newcommand{\evoskill}{EvoSkill}
\newcommand{\evoskillv}{EvoSkill~v2}
\newcommand{\skillopt}{SkillOpt}

\newcommand{\tauthree}{$\tau^3$-bench}
\newcommand{\claudecode}{Claude Code}

\title{\evoskillv{}: Fully Agentic Skill Evolution\\from Benchmark Adaptation to a Stateful Evolver}

\author{%
  \begin{tabular}{c@{\hspace{3em}}c@{\hspace{3em}}c}
    Dastin Huang & Baran Nama & Salaheddin Alzubi\\
    \authoremail{dastin@sentient.xyz} & \authoremail{baran@sentient.xyz} & \authoremail{salaheddin@sentient.xyz}\\[1.2ex]
    \multicolumn{3}{c}{Sentient Labs}
  \end{tabular}
}
\date{}

\usepackage[skip=4pt plus 1pt]{parskip}%
\begin{document}
\maketitle

\begin{abstract}
\evoskillv{} is a three-agent system that edits the instructions (\emph{skills}) an agent is given, leaving its model and harness intact. The user starts a run by describing a benchmark and supplying a grader, which a \emph{benchmark adaptor} turns into a benchmark-specific plugin. A \emph{solver} attempts the tasks, and an \emph{evolver} reads the solver's training rollouts and rewrites the skills at each epoch. We implement the evolver as one stateful coding agent with a bash tool for the whole run, unlike the two most common designs among the skill optimizers we surveyed: a sequence of single model calls, as in \skillopt{}, and an agent that starts afresh each round, as in SkillBoost. We also measure how much the solver's rollouts vary from one attempt to the next, give the evolver several attempts per training task, and accept every epoch's skills instead of selecting among them on noisy scores. With fine-grained ground truth, \evoskillv{}'s skills raised the held-out pass rate above the stronger of \skillopt{} and SkillBoost by $7.7$ percentage points on SpreadsheetBench~v2 and by $5.4$ on \tauthree{} banking. In both benchmarks and both feedback settings, \evoskillv{}'s evolver cost less per percentage point of held-out pass rate gained than either baseline's optimizer. \evoskillv{} also enables a less capable model to improve its own skills, demonstrated by a gain of up to $21.5$ percentage points in the \tauthree{} banking experiment where Claude Haiku~5.5 powers both solver and evolver.
\end{abstract}

\section{Introduction}
\label{sec:intro}

What an LLM agent does on a task is set as much by the content of the model's context window as by the model's weights: a system prompt, a set of tool descriptions, a folder of instruction files it opens when a
task matches the file's own description of when it applies. That context has become an optimization target in its own right.
GEPA~\cite{gepa2025} evolves the prompts of the model calls in a multi-step program, reflecting on the program's execution traces (its reasoning, tool calls and tool outputs); \skillopt{}~\cite{skillopt2026}
edits a single skill document (a short instruction file; defined below) of limited length, keeping an edit only if it passes a check on a validation set; a line of skill-library systems ---
Voyager~\cite{voyager2023}, ExpeL~\cite{expel2023}, Agent Workflow Memory~\cite{awm2024},
SkillWeaver~\cite{skillweaver2025}, ACE~\cite{ace2025} --- accumulates reusable procedures from an
agent's own experience. The appeal is that the optimized artifact is legible: unlike a weight
update, an instruction can be read, disputed, and given unchanged to a different agent. Our method builds on self-reflection, in which an agent reads its own failed attempt and tries again~\cite{shinn2023reflexion,madaan2023selfrefine}. In \evoskillv{}, a separate agent (the evolver) reads the step-by-step records of the solver's task attempts (its \emph{rollouts}) and updates the solver's skill set accordingly.

Earlier work has shown that such instructions can raise task success~\cite{gepa2025,ace2025,skillopt2026,lin2026rethinking,alzubi2026evoskill}; this paper asks how skill evolution is best done: what kind of optimizer writes the skills, what it remembers between rounds, how many attempts per task it sees, and how it decides which skills to keep (\S\ref{sec:design}).
The method rests on a narrow observation: \textbf{an agent without suitable skills that fails a task usually fails it again.} In our experiments, a Sonnet~5.5 Claude Code agent failed all three attempts on nine in ten of the SpreadsheetBench~v2 held-out tasks that it failed at least once. On \tauthree{} banking, a customer-service benchmark, a Haiku~5.5 Claude Code agent failed all three attempts on about six in ten such tasks (Figures~\ref{fig:heldout-ssb} and~\ref{fig:heldout-tau3}). Where failures repeat, some of them are caused by missing \emph{instructions} rather than by missing capability ---
and instructions are natural language, so another agent can write them. Anthropic introduced such instructions as \emph{skills}~\cite{anthropic2025skills} and released the format as an open standard, which other coding agents, among them OpenAI's Codex and GitHub Copilot, have adopted~\cite{agentskills2025}. We follow this definition: a skill is a folder holding a markdown file, optionally with scripts, whose short description tells the agent when to open it.

\evoskill{}~\cite{alzubi2026evoskill} is a skill evolution framework whose optimizer is two agents: a \emph{Proposer} that analyzes the solver's failures and proposes changes, and a \emph{Generator} that writes those proposals into skill files. \evoskillv{} merges the two into a single \textbf{evolver} and adds a \textbf{benchmark adaptor} (BA): an agent that turns a natural-language description of the benchmark (the user's \emph{instruction}) and the user's own grader into the run's fixed task list and the training material the evolver may read (\S\ref{sec:system}). \evoskillv{} is a \textbf{layer on top of whatever agent harness an engineer already runs} (e.g., Claude Code, Codex).

\Needspace{5\baselineskip}
\paragraph{Contributions.}
\begin{enumerate}[leftmargin=1.4em,itemsep=1pt,topsep=2pt]
  \item \textbf{A fully agentic skill-evolution system.} Three coding agents: a BA that turns a prose instruction and the user's grader into a frozen plug-in, a solver, and an evolver (\S\ref{sec:system}).
  \item \textbf{Four design choices for the evolver (\S\ref{sec:design}).} The evolver is (i)~an agent with a bash tool, not a sequence of single model calls; (ii)~one stateful session for the whole run, not a fresh start each round; (iii)~given several attempts per training task, because a single attempt is too noisy to tell a better skill from chance (Table~\ref{tab:mixed-tasks}); and (iv)~made to accept every epoch's skills, because a keep-if-better check on such noisy scores can reject a good edit (Table~\ref{tab:noise-gap}).
  \item \textbf{A comparison with two recent skill optimizers on two benchmarks.} With the same solver, Claude Sonnet~5.5 as every optimizer and each optimizer given its own fine-grained feedback, \evoskillv{}'s held-out pass rate beat the stronger of \skillopt{}~\cite{skillopt2026} and SkillBoost~\cite{lin2026rethinking} by $7.7$ percentage points on SpreadsheetBench~v2 and $5.4$ on \tauthree{} banking (Figures~\ref{fig:heldout-ssb} and~\ref{fig:heldout-tau3}). With pass or fail only, it matched \skillopt{} on SpreadsheetBench~v2 and led it on \tauthree{} banking, with an evolver that cost a fraction of \skillopt{}'s optimizer (Tables~\ref{tab:ssb-cost} and~\ref{tab:tau3-cost}).
  \item \textbf{Self-evolution of a less capable model.} With Claude Haiku~5.5 powering both solver and evolver, \evoskillv{} raised the held-out pass rate on \tauthree{} banking by up to $21.5$ percentage points (Figure~\ref{fig:heldout-tau3}).
\end{enumerate}

\section{Related Work}
\label{sec:related}

\emph{Single-call reflective} optimizers revise a text artifact, such as a prompt or a skill document, without giving the revising model tools: the proposal step reads execution traces (records of what the agent under optimization did on each task and how it was scored) and emits a new artifact directly, with no inner loop of self-critique before the proposal is final. GEPA~\cite{gepa2025} is the clearest example. It evolves prompts under Pareto selection (keeping every candidate prompt that is best on at least one training example), making one model call per prompt it updates, in which the model reads the traces and writes the new prompt; its merge operator, the step that combines two candidate prompts into one, is ordinary code rather than a model call. The distinction is about the \emph{proposal step}, not whether the optimizer iterates: every optimizer here has an outer loop of propose, evaluate, repeat. \skillopt{}~\cite{skillopt2026} sits between the single-call reflective optimizers and the agentic optimizers described next. It evolves a single skill document through small edits under a budget on how much each step may change, keeping an edit only if it strictly raises the validation score, but its update is a parallel proposal per small batch of examples, after which a model merges the proposals in stages, so one optimization step costs many calls rather than one. Agentic optimizers give tools to the model that proposes changes: Hone~\cite{hone2026} uses a command-line coding agent as the step in GEPA that proposes changes; SIA~\cite{sia2026} jointly evolves the agent's harness code and weights; \skillopt{}-Lite~\cite{skilloptlite2026} drives a production coding agent through a loop that repeatedly edits a prompt file and keeps or discards each edit by its score on a fixed validation split. Our first generation~\cite{alzubi2026evoskill} introduced the agentic skill-folder loop this report extends, including the rule that the ground truth is shown only to the optimizer, never to the solver. The line of work on skill libraries, among them Voyager~\cite{voyager2023}, ExpeL~\cite{expel2023}, Agent Workflow Memory~\cite{awm2024}, SkillWeaver~\cite{skillweaver2025}, and ACE~\cite{ace2025}, builds up a library of reusable procedures from the agent's own past runs, usually written by the same agent that solves the tasks, and kept without any check on held-out data; CoEvoSkills~\cite{coevoskills2026} evolves skill packages against a verifier that stands in for the real scorer and is itself evolved alongside the skills. Every system above is an \emph{optimizer}, and in each the measurement instrument (the split, the scorer, the models that run the tasks) is chosen outside the optimizing agent and cannot be changed by it: GEPA~\cite{gepa2025} takes its validation set, task model and metric as inputs from whoever runs it, and never changes them afterwards; SIA~\cite{sia2026} rewrites the source code of the agent being improved, starting from an initial version, while the evaluator and its ground truth stay in the task package. Setting the instrument outside the optimizing agent is a valid design, and we do not claim to improve on it. What differs here is that part of building the instrument is delegated to an agent (\S\ref{sec:system-ta}): the user supplies the grader and the data, and an agent turns the user's prose description into the task list, with each task's set, and into the material the evolver may read. Once an agent does that, it must not also be the agent that writes the skills, or a low score can be raised by changing the split (which examples count, and how many) or what the evolver may see, instead of by writing a better skill. None of the systems surveyed delegates this step to an agent, so none needs to separate that agent from the one that writes the skills. The failure modes of optimizing against a validation set are known, but the work on such optimizers rarely audits its own runs for them: the hazard of re-using one held-out set for many adaptive decisions is well studied~\cite{dwork2015holdout} and shortcut learning, in which a model picks up spurious cues that predict the label without solving the task, is documented for trained models~\cite{geirhos2020shortcut,gururangan2018annotation}, but here the \emph{optimizer} is the learner. It can write rules that fit the training tasks and do not carry over to other tasks: in our SpreadsheetBench~v2 run and our \tauthree{} run without training answers, the training scores rose far more than the validation scores (\S\ref{sec:results-ssb}, \S\ref{sec:results-tau3}). It can also find quirks in how the task was built, including quirks of the grader itself (\S\ref{sec:discussion-grader}). Huang et al.~\cite{skilllens2026} find that skills extracted by strong models do not reliably benefit less capable models; in this report the evolved skills were evaluated only with the solver model they were written for.

\section{The \evoskillv{} System}
\label{sec:system}

\begin{figure*}[!ht]
\centering
\includegraphics[width=\textwidth]{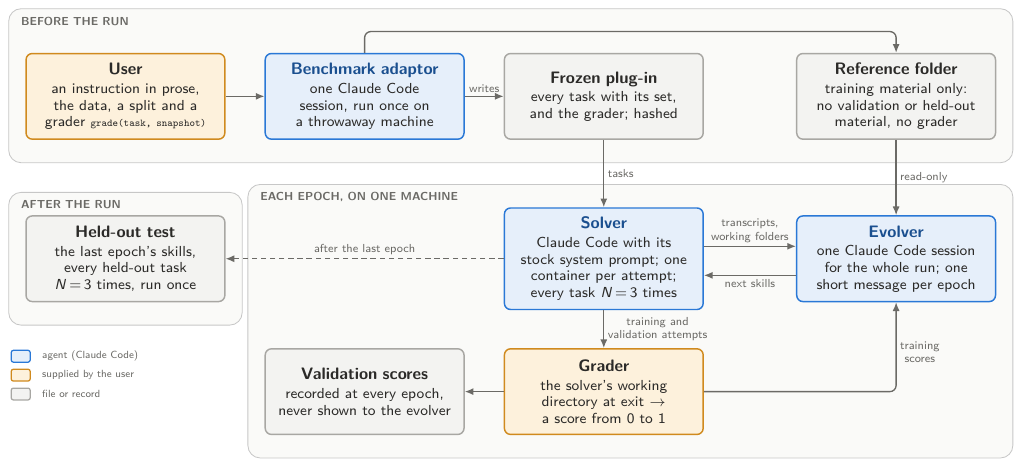}%
\caption{\evoskillv{}. Before the run, the benchmark adaptor (BA) turns the user's instruction, data, split and grader into a frozen plug-in that lists every task with its set, and a reference folder of training material. In each epoch the solver attempts every training and validation task three times, and the user's grader scores each attempt. The evolver, one Claude Code session for the whole run, reads only the training attempts, their scores and the reference folder, and writes the next skills. The last epoch's skills are tested on the held-out set.}
\label{fig:architecture}
\end{figure*}

\evoskillv{} is three coding agents layered over an existing agent harness: a \textbf{benchmark adaptor} (\textbf{BA}) that turns the user's description of a benchmark into the run's fixed list of tasks before any skill is written; a \textbf{solver} that attempts the tasks with the evolved skills; and an \textbf{evolver} that reads the solver's attempts and rewrites the skills. In our implementation, all three are Claude Code sessions (Figure~\ref{fig:architecture}).

\subsection{The Benchmark Adaptor}
\label{sec:system-ta}

\textbf{In v1, a new benchmark had to fit one of the input formats it supports.} A benchmark whose output is a file or a conversation, such as SpreadsheetBench~v2's workbooks or \tauthree{}'s customer dialogues, first had to be converted by hand into one of them, and v1 has limited support for custom graders.

\textbf{BA is introduced in v2 to automate the per-benchmark adaptation.} The user hands over three things: a short instruction in natural language (the benchmark, where its data is, the split to use or how to make one, what the solver gets and must produce, and the run size), the dataset, and a grader. The grader is a function \texttt{grade(task, snapshot)}: it takes the solver's working directory as it was at the end of an attempt and returns a score from $0$ to $1$. The BA generates two things. The first is a \emph{plug-in}: a short Python module whose one function lists every task with its set (train, validation or test) and with the fields that the user's grader reads. The second is a \emph{reference folder}: everything the user handed over that the evolver may read, which is everything except material about validation or test tasks and the grader itself.

\subsection{The solver}
\label{sec:system-solver}

The solver is a Claude Code agent with its stock system prompt. The task reaches it as the first user message, and the solver works in its own container. A \emph{skill} is a folder holding a \texttt{SKILL.md} file and, optionally, scripts. Skills reach the solver by \emph{progressive disclosure}~\cite{anthropic2025skills}: Claude Code keeps every skill's name and one-line description in the solver's context at all times, and the solver loads the full body by a tool call only when that description matches the task in front of it. So the descriptions, which decide when a skill is loaded, are themselves evolvable. In the runs reported here each attempt had $1$\,GiB of memory; web search, web fetch and subagents were disabled; and the container could reach no network host except the model API.

\subsection{The evolver, and what it is given}
\label{sec:system-evolver}

The evolver is a single Claude Code session that lasts the whole run, in a container on the same machine as the solver's containers. It gets Claude Code's own system prompt with a fixed text appended once and one short message per epoch to analyze the solver's rollouts, diagnose the pitfalls that a skill could fix, and write the next skill set into a staging folder (both texts in Appendices~\ref{sec:appendix-evolver-system} and~\ref{sec:appendix-evolver-message}). Because the session continues from one epoch to the next, the evolver keeps its own earlier analysis in its context. Claude Code may compact that context when it grows too large, so the evolver also has a private folder, kept across epochs, for anything it wants to keep. Table~\ref{tab:evolver-view} lists what it finds on its machine.

\begin{table}[!ht]
\centering\small
\begin{tabular}{@{}lp{0.56\linewidth}l@{}}
\toprule
Path on the evolver's machine & What it holds & Access \\
\midrule
\texttt{/evo/traces/epoch\_$K$/} & each training attempt's Claude Code session log, unchanged, and the large tool outputs it saved to files & read \\
\texttt{/evo/scores/epoch\_$K$.json} & each training attempt's score ($0$ to $1$) or \emph{grader\_failed} & read \\
\texttt{/evo/solver\_workspaces/epoch\_$K$/} & the solver's working directory at the end of each training attempt & read \\
\texttt{/evo/skills/epoch\_$K$/} & the skill set that epoch $K$'s solver carried & read \\
\texttt{/evo/reference/} & the BA's reference folder (training material only) & read \\
\texttt{/evo/staging/} & the next skill set; it starts as the set the evolver left there & write \\
\texttt{/evo/evolver\_workspace/} & the evolver's private folder, kept across epochs & write \\
\bottomrule
\end{tabular}
\caption{What the evolver finds on its machine; epoch~$0$ ran with no skills. What never reaches it: \S\ref{sec:system-isolation}.}
\label{tab:evolver-view}
\end{table}

\subsection{Keeping held-out tasks away from the evolver}
\label{sec:system-isolation}

\textbf{A common pitfall in self-improving systems is that the optimizer reads the test split directly and overfits to it} --- or worse, manipulates the split itself to drop the tasks it fails on. Either one raises the score without improving a single skill, and neither leaves a mark: the run looks exactly like a run that worked. In \evoskillv{} the split is fixed in the frozen plug-in before the first skill is written, and the evolver's machine receives only training material: the reference folder, and the solver's training attempts with their scores. Validation and test attempts, and their scores, are never uploaded to it. This isolation is set by the orchestration layer, so it holds even if the evolver tries to work around it.

\textbf{The isolation ensures that a score obtained with an evolved skill was not obtained by reading held-out answers} --- but a higher score does not necessarily mean the skill is a good one. When the scoring mechanism itself is flawed, the evolver may optimize in a direction the user did not intend, even though the isolation works exactly as designed.

\paragraph{A grader defect that an earlier version found.}
\phantomsection
\label{sec:discussion-grader}
An earlier version of this system, set to maximize the SpreadsheetBench~v2 score, found a defect in the
benchmark's grader and wrote a skill to exploit it. The grader compared the value that a workbook stores beside each formula, as the spreadsheet
application last computed it, with the expected workbook, and did not recompute the formulas. The evolver wrote a skill
that told the solver to repair the formula text and never to recalculate~\cite{evoskillblog2026}; the grader used here recalculates every formula (Appendix~\ref{sec:appendix-ssb-split-grader}). An optimizer can find a defect like this one, so the grader needs checking before the run; this is
one reason why the user, not an agent, supplies it.

\section{Experimental Setup}
\label{sec:setup}

We use two benchmarks (\S\ref{sec:setup-ssb}, \S\ref{sec:setup-tau3}), chosen because both are hard: the best scores on their public leaderboards are $54.5\%$ on SpreadsheetBench~v2~\cite{ssb2leaderboard} and $55.2\%$ on \tauthree{} banking~\cite{taubenchleaderboard}, which leaves substantial room for improvement. Both are also all-or-nothing: an \emph{episode}, one attempt at one task, scores $1$ only when every check passes. A skill therefore counts only if it is solid enough to make the solver pass a task that it fails without the skill; making an attempt partly right earns nothing.

For each benchmark we wrote the BA's user instruction by hand (Appendices~\ref{sec:appendix-instruction-ssb} and~\ref{sec:appendix-instruction-tau3}). The BA and the evolver are Claude Code agents powered by Claude Sonnet~5.5, except in two \tauthree{} runs whose evolver is powered by the solver's own model, Claude Haiku~5.5 (\S\ref{sec:results-tau3}).

\label{sec:setup-current}

\subsection{SpreadsheetBench~v2}
\label{sec:setup-ssb}

In SpreadsheetBench~v2~\cite{zhu2026spreadsheetbench2}, the solver, a Claude Code agent powered by Claude Sonnet~5.5, edits a real Excel workbook, and the result is compared cell by cell with a reference workbook; our grader first recalculates every formula in LibreOffice (\S\ref{sec:discussion-grader}). We use three of the benchmark's four categories, Debugging, Financial\_Model and Template ($297$ tasks); the fourth, Visualization, is the least challenging of the four in the benchmark's own evaluation~\cite{zhu2026spreadsheetbench2}, and its grading needs a vision-language model. Some tasks could not be used as given, for example because a reference workbook's saved values differ from a fresh recalculation. After filtering, $218$ tasks remain, split into $57$ training, $57$ validation and $104$ held-out tasks.

Tasks that share a reference workbook, a formula family or a Template folder go into the same set, so no answer is shared across sets. Each attempt has a $20$-minute wall-clock limit and a $45$-turn limit. Appendix~\ref{sec:appendix-ssb-split-grader} gives the split and the grader in full.

In the \evoskillv{} runs, we give the BA a short user instruction that describes SpreadsheetBench~v2 (Appendix~\ref{sec:appendix-instruction-ssb}) and the benchmark's dataset as downloaded from its Hugging Face release~\cite{ssb2dataset}. The BA is instructed to keep the grader and everything about the validation and held-out tasks out of the evolver's reference folder (Appendix~\ref{sec:appendix-ba-guidance}), and we checked by hand that it did. The evolver may read everything else in that folder, including the training tasks' expected workbooks. \skillopt{}'s and SkillBoost's optimizers read each training attempt's session log and whether it passed. For each failed attempt they also get the wrong cells, with the expected and actual values, as the official implementations of both methods provide: \skillopt{} every wrong cell, SkillBoost the first. Appendix~\ref{sec:appendix-ssb-baselines} lists where our runs of the two baselines differ from their released code.

\subsection{\texorpdfstring{\tauthree{}}{tau3-bench} banking}
\label{sec:setup-tau3}

In \tauthree{} banking~\cite{shi2026tauknowledge}, the solver, a Claude Code agent powered by Claude Haiku~5.5, plays a bank's customer-service agent talking with a simulated customer. Each task is graded on \emph{database state} (the bank's final database must equal the one that the expected tool calls produce) or on \emph{action} (the agent must make each expected call with the expected arguments).

After data cleansing, we split the remaining $70$ tasks into $19$ training, $20$ validation and $31$ held-out tasks, keeping related tasks in one set.

In the \evoskillv{} runs, we give the BA a short user instruction that describes \tauthree{} banking (Appendix~\ref{sec:appendix-instruction-tau3}) and the benchmark's banking tasks and the files they use, from its public repository~\cite{tau2benchrepo}. As on SpreadsheetBench~v2, the BA keeps the grader and everything about the validation and held-out tasks out of the evolver's reference folder, and we checked that it did. The evolver may read everything else. \skillopt{}'s and SkillBoost's optimizers read each training attempt's conversation and whether it passed. For each failed attempt they also get what the grader found wrong. \skillopt{} gets every difference, SkillBoost the first.

\label{sec:setup-taxonomy}

\section{Results}
\label{sec:results}

\subsection{SpreadsheetBench~v2}
\label{sec:results-ssb}
\label{sec:results-baselines}

\begin{figure*}[!ht]%
\centering
\includegraphics[width=\linewidth]{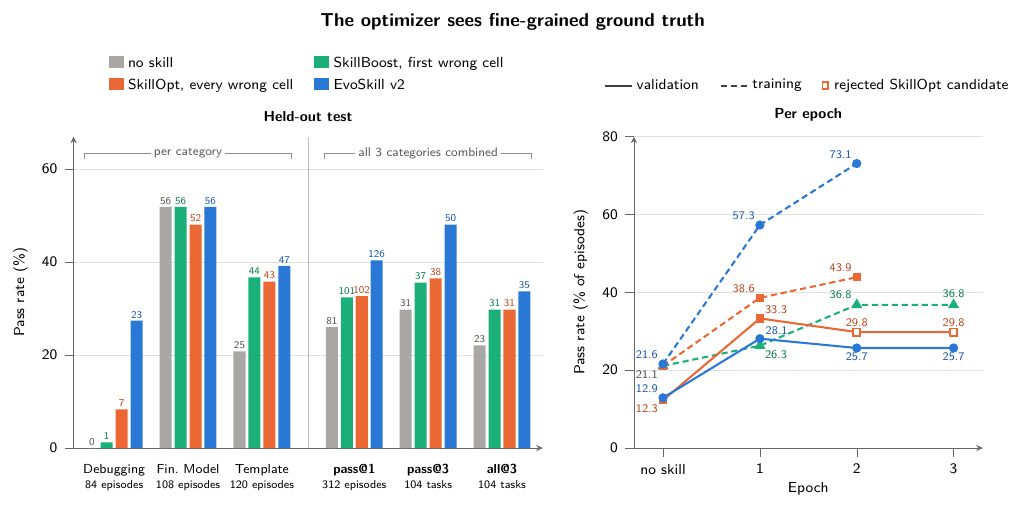}
\caption{SpreadsheetBench~v2, the runs whose optimizer sees fine-grained ground truth (\S\ref{sec:setup-ssb}). Bars: share of episodes passed (pass@1), and of tasks passed in at least one of $3$ attempts (pass@3) and in all three (all@3); the number above a bar is the count. Per epoch: each \evoskillv{} point is of $171$ episodes, each baseline point of $57$; SkillBoost has no validation score, and its flat segments are rounds that kept no candidate; grey labels are values two lines share.}
\label{fig:heldout-ssb}
\end{figure*}

We compare \evoskillv{} with \skillopt{}~\cite{skillopt2026} and SkillBoost~\cite{lin2026rethinking} on the same split. We keep the solver configuration, data split and grading metrics identical across $3$ experiments. We run the solver $3$ times, independently, on each of the $104$ held-out tasks for each skill. With \evoskillv{}'s skills the solver passed $126$ of $312$ held-out episodes, against at most $102$ with the skills of either baseline (Figure~\ref{fig:heldout-ssb}, left; Appendix~\ref{sec:appendix-ssb-skills} describes \evoskillv{}'s skills).

\begin{figure*}[!ht]%
\centering
\includegraphics[width=\linewidth]{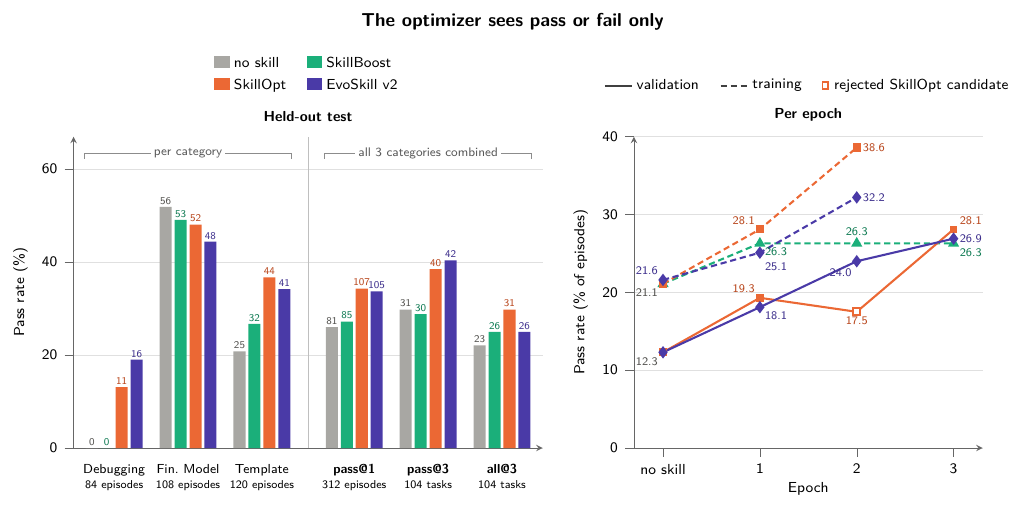}
\caption{SpreadsheetBench~v2, the runs whose optimizer sees only whether each training attempt passed; \evoskillv{}'s evolver reads the BA's reference folder without the training tasks' expected workbooks. Panels as in Figure~\ref{fig:heldout-ssb}; \evoskillv{} ran with its own no-skill training and validation legs.}
\label{fig:ssb-pf}
\end{figure*}

\paragraph{Sparse feedback signal.}
In the first experiment, \evoskillv{}'s evolver has far more ground truth for diagnosing the solver's failures than \skillopt{}'s and SkillBoost's optimizers, so it is hard to tell whether \evoskillv{}'s higher score comes from its evolution loop itself. We therefore ran a second experiment that changes what every optimizer sees: only whether each training attempt passed (Figure~\ref{fig:ssb-pf}). With only this sparse signal, \evoskillv{} is on par with \skillopt{} on the held-out set at less than a tenth of \skillopt{}'s optimizer cost, and SkillBoost passed fewer episodes at the highest optimizer cost (Table~\ref{tab:ssb-cost}).

\begin{table}[!ht]%
\centering\small\setlength{\tabcolsep}{4.5pt}
\begin{tabular}{@{}lrrrrrr@{}}
\toprule
 & \multicolumn{3}{c}{Fine-grained ground truth} & \multicolumn{3}{c}{Pass or fail only} \\
\cmidrule(lr){2-4}\cmidrule(l){5-7}
 & Held-out & Cost & Per point gained & Held-out & Cost & Per point gained \\
\midrule
No skill & 81 & --- & --- & 81 & --- & --- \\
SkillBoost & 101 & \$13.95 & \$2.18 & 85 & \$13.66 & \$10.65 \\
\skillopt{} & 102 & \textbf{\$8.56} & \$1.27 & \textbf{107} & \$6.65 & \$0.80 \\
\evoskillv{} & \textbf{126} & \$12.84 & \textbf{\$0.89} & 105 & \textbf{\$0.53} & \textbf{\$0.07} \\
\bottomrule
\end{tabular}
\caption{SpreadsheetBench~v2: held-out episodes passed (of $312$) and the optimizer's cost, in Claude Code's own list-price estimates for the optimizer alone; the solver's attempts are not included. Per point gained: the cost divided by the percentage points of held-out pass rate gained over no skill. The two \evoskillv{} runs used different Claude Code versions, so their costs are not directly comparable. Bold marks each column's best: the highest score or the lowest cost.}
\label{tab:ssb-cost}
\end{table}

\subsection{\texorpdfstring{\tauthree{}}{tau3-bench} banking}
\label{sec:results-tau3}

\begin{figure*}[!ht]%
\centering
\includegraphics[width=0.95\linewidth]{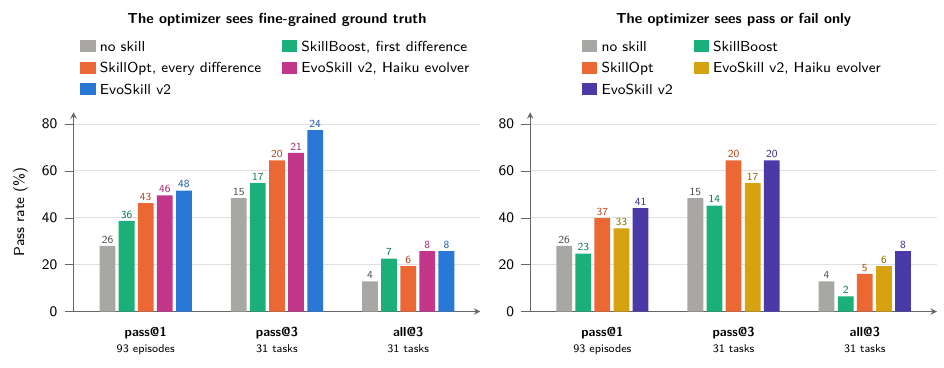}
\caption{\tauthree{} banking held-out test, for the runs whose optimizer sees fine-grained ground truth (\S\ref{sec:setup-tau3}) and those whose optimizer sees only whether each training attempt passed. Bars as in Figure~\ref{fig:heldout-ssb}.}
\label{fig:heldout-tau3}
\end{figure*}

We compare the same three methods on \tauthree{} banking, with Claude Haiku~5.5 as the solver. As on SpreadsheetBench~v2, the solver, its limits per attempt, the customer, the grader and the split are the same for every run. With fine-grained ground truth, where \evoskillv{}'s evolver reads the $19$ training task files with each task's customer script and expected tool calls (the \emph{training answers}), \evoskillv{}'s skills passed $48$ of $93$ held-out episodes, against at most $43$ with the skills of either baseline (Figure~\ref{fig:heldout-tau3}, left; Appendix~\ref{sec:appendix-tau3-skills} describes \evoskillv{}'s skills).

\paragraph{Sparse feedback signal.}
With only pass or fail, \evoskillv{}'s skills again passed the most held-out episodes (Figure~\ref{fig:heldout-tau3}, right), and in both settings its evolver cost far less than either baseline's optimizer (Table~\ref{tab:tau3-cost}).

\begin{table}[!ht]%
\centering\small\setlength{\tabcolsep}{4.5pt}
\begin{tabular}{@{}lrrrrrr@{}}
\toprule
 & \multicolumn{3}{c}{Fine-grained ground truth} & \multicolumn{3}{c}{Pass or fail only} \\
\cmidrule(lr){2-4}\cmidrule(l){5-7}
 & Held-out & Cost & Per point gained & Held-out & Cost & Per point gained \\
\midrule
No skill & 26 & --- & --- & 26 & --- & --- \\
SkillBoost & 36 & \$6.98 & \$0.65 & 23 & \$8.64 & no gain \\
\skillopt{} & 43 & \$5.55 & \$0.30 & 37 & \$6.22 & \$0.53 \\
\evoskillv{}, Haiku evolver & 46 & \$1.87 & \$0.09 & 33 & \textbf{\$0.73} & \textbf{\$0.10} \\
\evoskillv{} & \textbf{48} & \textbf{\$0.99} & \textbf{\$0.04} & \textbf{41} & \$1.55 & \textbf{\$0.10} \\
\bottomrule
\end{tabular}
\caption{\tauthree{} banking: held-out episodes passed (of $93$) and the optimizer's cost, as in Table~\ref{tab:ssb-cost}; here every run used the same Claude Code version. Bold marks each column's best; ``no gain'': SkillBoost's skills passed fewer held-out episodes than no skill.}
\label{tab:tau3-cost}
\end{table}

\begin{figure}[!ht]
\centering
\includegraphics[width=\linewidth]{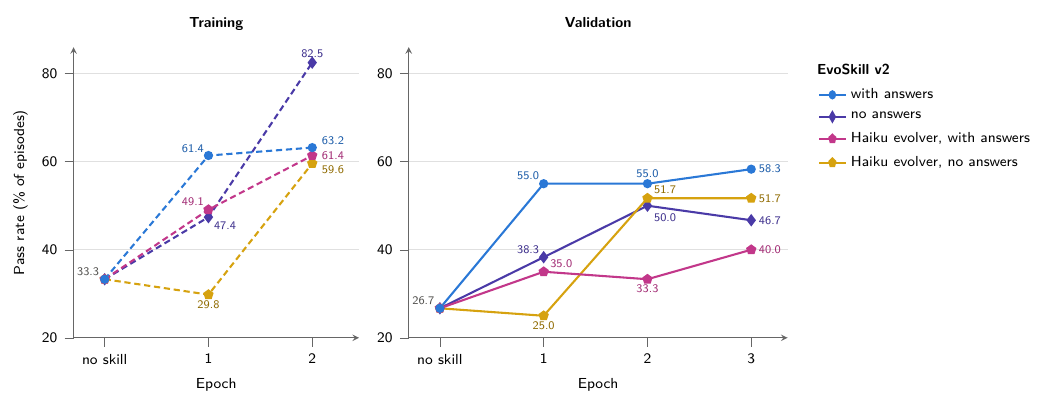}%
\caption{\tauthree{} banking: \evoskillv{}'s pass rate per epoch on the training (left) and validation (right) tasks, each point of $57$ or $60$ episodes ($3$ attempts on $19$ training or $20$ validation tasks), with Sonnet~5.5 or Haiku~5.5 as its evolver, each with and without the training answers (\S\ref{sec:results-tau3}).}
\label{fig:epochs-tau3}
\end{figure}

\paragraph{Self-evolution.}
On \tauthree{} banking, every run above powers the evolver with a stronger model than the solver's (Sonnet~5.5 against Haiku~5.5), so the held-out gain combines a better evolver model with the \evoskillv{} harness. To separate the two, and to test whether the harness lets a model improve its own skills, we ran a third experiment in which the solver's own model, Claude Haiku~5.5, powers the evolver, with each of the two feedback signals. With fine-grained ground truth, its skills came within two held-out episodes of the Sonnet~5.5 evolver's; with pass or fail only, they gained about half as much over no skill (Figure~\ref{fig:heldout-tau3} and Table~\ref{tab:tau3-cost}). Figure~\ref{fig:epochs-tau3} shows the four \evoskillv{} runs per epoch.

\section{Discussion: Four Design Choices}
\label{sec:design}

With fine-grained ground truth, \evoskillv{}'s skills passed more held-out episodes than either baseline's on both benchmarks (Figures~\ref{fig:heldout-ssb} and~\ref{fig:heldout-tau3}). Four choices in \evoskillv{}'s design, the same on both benchmarks, set it apart from both baselines.

\subsection{The evolver is an agent with a bash tool, not a sequence of single model calls}
\label{sec:design-agent}

Most recent skill optimizers, among them \skillopt{}, GEPA~\cite{gepa2025} and ACE~\cite{ace2025}, edit skills through single model calls, each answering from the text in its prompt. In the SpreadsheetBench~v2 run with fine-grained ground truth, \skillopt{} made $40$ such calls: most analyze a batch of training attempts, and the rest merge or rank their proposals, revise the skill, or write notes for later calls. None of them can open a workbook, recalculate a formula or check a conclusion before proposing an edit. We argue that an agent suits the job better, because it can check an explanation of a failure before it writes that explanation into a skill; SkillOpt-Lite~\cite{skilloptlite2026} and CoEvoSkills~\cite{coevoskills2026} also give their optimizer a shell.

\evoskillv{}'s evolver is a \claudecode{} agent with a bash tool. In our experiments it used bash mostly to run Python, and in some of its turns it also handed groups of training tasks to subagents with bash tools of their own. On SpreadsheetBench~v2 with fine-grained ground truth, one script it wrote recalculated a solver's output workbook in LibreOffice and compared it cell by cell with the training task's expected workbook, so the evolver could test an explanation of a failure before writing it into a skill.

\subsection{The evolver is stateful: one session for the whole run}
\label{sec:design-state}

SkillBoost starts a new optimizer session every round, and each of \skillopt{}'s calls starts from its own prompt, so what an earlier round or call worked out reaches a later one only as text it passed on, such as the skill or \skillopt{}'s notes. Among the agent-based optimizers we surveyed that run in rounds, starting a new session each round is the more common design ($8$, against $5$ that keep one session). \evoskillv{}'s evolver is one \claudecode{} session for the whole run: in the SpreadsheetBench~v2 run with fine-grained ground truth, its three turns continued one session, whose context \claudecode{} summarized four times when it filled. At each turn the evolver has every earlier epoch's training attempts and scores, every skill set it wrote before (Table~\ref{tab:evolver-view}), its notes, its own earlier reasoning and the scripts in its private folder; in its second and third turns, $20$ of its bash commands re-ran scripts it had written in an earlier turn. Each turn is therefore a chance to keep, revert or combine anything tried so far, not only to edit the latest version.

\subsection{Several attempts per task make the evolver's evidence less noisy}
\label{sec:design-resampling}

The solver does not pass or fail a task the same way every time. In every training and validation leg, some tasks passed some of their three attempts and failed others: $7\%$ to $21\%$ of them on SpreadsheetBench~v2 and $16\%$ to $53\%$ on \tauthree{} banking (Table~\ref{tab:mixed-tasks}). With one attempt per task, as \skillopt{}, SkillBoost and $35$ of the other $43$ skill optimizers we surveyed run, each of those tasks appears as a pass or a fail by chance. With three, the evolver can tell a failure that repeats from one that does not.

\begin{table}[!ht]%
\centering\footnotesize
\begin{minipage}[t]{0.42\linewidth}\centering\setlength{\tabcolsep}{5.2pt}
\begin{tabular}{@{}lrrrrrr@{}}
\toprule
 & \multicolumn{3}{c}{Training} & \multicolumn{3}{c}{Validation} \\
\cmidrule(lr){2-4}\cmidrule(l){5-7}
Pass rate & 3/3 & 0/3 & 1--2/3 & 3/3 & 0/3 & 1--2/3 \\
\midrule
\multicolumn{7}{@{}l}{\emph{SpreadsheetBench~v2 ($57$ tasks per set)}} \\
No skill & $10$ & $43$ & $4$ & $5$ & $47$ & $5$ \\
Turn~$1$ & $27$ & $19$ & $11$ & $14$ & $38$ & $5$ \\
Turn~$2$ & $36$ & $9$ & $12$ & $10$ & $38$ & $9$ \\
Turn~$3$ & --- & --- & --- & $11$ & $38$ & $8$ \\
\midrule
\multicolumn{7}{@{}l}{\emph{\tauthree{} banking ($19$ training, $20$ validation tasks)}} \\
No skill & $3$ & $9$ & $7$ & $3$ & $12$ & $5$ \\
Turn~$1$ & $10$ & $6$ & $3$ & $7$ & $5$ & $8$ \\
Turn~$2$ & $7$ & $2$ & $10$ & $6$ & $6$ & $8$ \\
Turn~$3$ & --- & --- & --- & $7$ & $4$ & $9$ \\
\bottomrule
\end{tabular}
\caption{Tasks by how many of their three attempts passed, with fine-grained ground truth (on \tauthree{}, the Sonnet~5.5 evolver). Turn~$k$: the skills of the evolver's $k$-th turn; ---: not run.}
\label{tab:mixed-tasks}
\end{minipage}\hfill
\begin{minipage}[t]{0.55\linewidth}\centering\setlength{\tabcolsep}{4.5pt}
\begin{tabular}{@{}llll@{}}
\toprule
 & Set & Passed & Newly failed \\
\midrule
\multicolumn{4}{@{}l}{\emph{SpreadsheetBench~v2 ($57$ tasks per set)}} \\
\skillopt{}, step~$2$ & Validation & $17$ against $19$ & --- \\
\skillopt{}, step~$3$ & Validation & $17$ against $19$ & --- \\
\skillopt{}, final skill & Validation & $18$ against $19$ & --- \\
SkillBoost, round~$3$ & Training & $20$ and $21$ against $21$ & $4$ and $2$ \\
\midrule
\multicolumn{4}{@{}l}{\emph{\tauthree{} banking ($19$ training, $20$ validation tasks)}} \\
\skillopt{}, step~$2$ & Validation & $9$ against $9$ & --- \\
\skillopt{}, step~$3$ & Validation & $8$ against $9$ & --- \\
\skillopt{}, final skill & Validation & $8$ against $9$ & --- \\
SkillBoost, round~$1$ & Training & $7$ and $9$ against $5$ & $1$ and $1$ \\
SkillBoost, round~$3$ & Training & $10$ and $7$ against $8$ & $1$ and $3$ \\
\bottomrule
\end{tabular}
\caption{Candidates that \skillopt{} and SkillBoost rejected in the runs with fine-grained ground truth, each scored with one attempt per task against the skill held at the time. \skillopt{} keeps a candidate only if it passes more validation tasks; SkillBoost only if it passes more training tasks and newly fails fewer than $5\%$ of them.}
\label{tab:noise-gap}
\end{minipage}
\end{table}

\subsection{Every epoch's skills are accepted}
\label{sec:design-accept}

Most skill optimizers keep a new candidate skill only if it passes a check on its scores: \skillopt{} requires a higher validation score, and SkillBoost requires more training tasks passed with few newly failed. With one attempt per task, the sampling noise discussed in \S\ref{sec:design-resampling} could easily flip a decision on whether or not keep a candidate. In the runs with fine-grained ground truth, every rejection by \skillopt{} and SkillBoost turned on $2$ tasks or fewer (Table~\ref{tab:noise-gap}), fewer than the tasks whose outcome varies between attempts (Table~\ref{tab:mixed-tasks}). In other words, had the attempts been sampled again, any of these rejected candidates could have been kept.

Several attempts per task reduce this noise but cannot remove it, so any rule that keeps or rejects a skill set by its rollout scores stays exposed to it. \evoskillv{} has no rejection step: it keeps every skill set the evolver writes, and tests the last epoch's on the held-out set instead of choosing an epoch by a validation score the evolver never sees. A bad edit is not lost. The evolver keeps every earlier epoch's skill sets and rollouts (\S\ref{sec:design-state}), so the next epoch's training attempts show what the edit made worse, and a later turn can revert it: the failed edit becomes evidence instead of being thrown away.

The last epoch's skills are not always the best on validation, but choosing the epoch with the best validation score would not have done better overall. In all six \evoskillv{} runs, validation rose from no skill to the last epoch (Figures~\ref{fig:heldout-ssb}, \ref{fig:ssb-pf} and~\ref{fig:epochs-tau3}). In four of them the last epoch also scored highest or tied, so choosing by validation would have kept the same skills. It would have chosen an earlier epoch only in the other two, and there the earlier epoch's skills passed $15$ fewer held-out episodes on SpreadsheetBench~v2 with fine-grained ground truth ($111$ against $126$ of $312$) and one more on \tauthree{} banking.

\section{Conclusion}
\label{sec:conclusion}

\evoskillv{} layers three coding agents (a benchmark adaptor, a solver, and an evolver that rewrites the solver's skills) on top of an existing harness. Its evolver is one stateful agent with a bash tool that sees several attempts per training task. With fine-grained ground truth, its skills passed more held-out episodes than those of \skillopt{} and SkillBoost on both SpreadsheetBench~v2 and \tauthree{} banking; with pass or fail only, it was on par with \skillopt{} on SpreadsheetBench~v2 and ahead of it on \tauthree{} banking, at a fraction of \skillopt{}'s optimizer cost; and it let Claude Haiku~5.5 improve its own skills on \tauthree{} banking. \evoskillv{} tests the last epoch's skills; in our runs, choosing among epochs by validation score would not have done better overall (\S\ref{sec:design-accept}).

\label{main-end}%

\FloatBarrier
\bibliographystyle{plainnat}
\bibliography{refs}

\begin{thebibliography}{28}
\providecommand{\natexlab}[1]{#1}
\providecommand{\url}[1]{\texttt{#1}}
\expandafter\ifx\csname urlstyle\endcsname\relax
  \providecommand{\doi}[1]{doi: #1}\else
  \providecommand{\doi}{doi: \begingroup \urlstyle{rm}\Url}\fi

\bibitem[{Agent Skills}(2025)]{agentskills2025}
{Agent Skills}.
\newblock Agent skills: an open format for giving agents new capabilities.
\newblock \url{https://agentskills.io}; specification at
  \url{https://github.com/agentskills/agentskills}, 2025.
\newblock Accessed 16 October 2026.

\bibitem[Agrawal et~al.(2025)Agrawal, Tan, Soylu, Ziems, Khare, Opsahl-Ong,
  Singhvi, Shandilya, Ryan, Jiang, Potts, Sen, Dimakis, Stoica, Klein, Zaharia,
  and Khattab]{gepa2025}
Lakshya~A Agrawal, Shangyin Tan, Dilara Soylu, Noah Ziems, Rishi Khare, Krista
  Opsahl-Ong, Arnav Singhvi, Herumb Shandilya, Michael~J Ryan, Meng Jiang,
  Christopher Potts, Koushik Sen, Alexandros~G Dimakis, Ion Stoica, Dan Klein,
  Matei Zaharia, and Omar Khattab.
\newblock Gepa: Reflective prompt evolution can outperform reinforcement
  learning.
\newblock \emph{arXiv preprint arXiv:2507.19457}, 2025.

\bibitem[Alzubi et~al.(2026{\natexlab{a}})Alzubi, Nama, Huang, and
  Saxena]{evoskillblog2026}
Salaheddin Alzubi, Baran Nama, Dastin Huang, and Abhishek Saxena.
\newblock {Dario Amodei warned about AI gaming its grader. Here's what happened
  when ours did}.
\newblock Sentient blog,
  \url{https://www.sentient.xyz/blog/self-evolving-agent-evoskill-v2},
  2026{\natexlab{a}}.
\newblock 16 September 2026.

\bibitem[Alzubi et~al.(2026{\natexlab{b}})Alzubi, Provenzano, Calvo, Bingham,
  Chen, Tyagi, Huang, and Vu]{alzubi2026evoskill}
Salaheddin Alzubi, Noah Provenzano, Christian~Alexander Calvo, Jaydon Bingham,
  Weiyuan Chen, Himanshu Tyagi, Dastin Huang, and Tu~Vu.
\newblock Evoskill: Automated skill discovery for agents.
\newblock In \emph{The First Workshop on Agent Skills (Agent Skills '26) at the
  ACM Conference on AI and Agentic Systems (CAIS 2026)}, 2026{\natexlab{b}}.
\newblock URL \url{https://openreview.net/forum?id=i4b4As7RFX}.

\bibitem[{Anthropic}(2025)]{anthropic2025skills}
{Anthropic}.
\newblock Agent skills.
\newblock
  \url{https://docs.claude.com/en/docs/agents-and-tools/agent-skills/overview},
  2025.
\newblock Claude platform documentation; Agent Skills announced 16 October
  2025.

\bibitem[Dwork et~al.(2015)Dwork, Feldman, Hardt, Pitassi, Reingold, and
  Roth]{dwork2015holdout}
Cynthia Dwork, Vitaly Feldman, Moritz Hardt, Toniann Pitassi, Omer Reingold,
  and Aaron Roth.
\newblock The reusable holdout: Preserving validity in adaptive data analysis.
\newblock \emph{Science}, 349\penalty0 (6248):\penalty0 636--638, 2015.

\bibitem[Geirhos et~al.(2020)Geirhos, Jacobsen, Michaelis, Zemel, Brendel,
  Bethge, and Wichmann]{geirhos2020shortcut}
Robert Geirhos, J{\"o}rn-Henrik Jacobsen, Claudio Michaelis, Richard Zemel,
  Wieland Brendel, Matthias Bethge, and Felix~A Wichmann.
\newblock Shortcut learning in deep neural networks.
\newblock \emph{Nature Machine Intelligence}, 2\penalty0 (11):\penalty0
  665--673, 2020.

\bibitem[Gururangan et~al.(2018)Gururangan, Swayamdipta, Levy, Schwartz,
  Bowman, and Smith]{gururangan2018annotation}
Suchin Gururangan, Swabha Swayamdipta, Omer Levy, Roy Schwartz, Samuel~R
  Bowman, and Noah~A Smith.
\newblock Annotation artifacts in natural language inference data.
\newblock In \emph{NAACL}, 2018.

\bibitem[Hebbar et~al.(2026)Hebbar, Manawat, Verboomen, Ivanova, Palanimalai,
  Bhatia, and Baskaran]{sia2026}
Prannay Hebbar, Yogendra Manawat, Samuel Verboomen, Alesia Ivanova, Selvam
  Palanimalai, Kunal Bhatia, and Vignesh Baskaran.
\newblock Sia: Self improving ai with harness \& weight updates.
\newblock \emph{arXiv preprint arXiv:2605.27276}, 2026.

\bibitem[Huang et~al.(2026)Huang, Xu, Yang, Gong, Yang, Tian, Wang, Lv, Gao,
  Dai, Liu, Qiu, Yang, Chen, Zheng, and Luo]{skilllens2026}
Zisu Huang, Jingwen Xu, Yifan Yang, Ziyang Gong, Qihao Yang, Muzhao Tian,
  Xiaohua Wang, Changze Lv, Xuemei Gao, Qi~Dai, Bei Liu, Kai Qiu, Xue Yang,
  Dongdong Chen, Xiaoqing Zheng, and Chong Luo.
\newblock From raw experience to skill consumption: A systematic study of
  model-generated agent skills.
\newblock \emph{arXiv preprint arXiv:2605.23899}, 2026.

\bibitem[Lin et~al.(2026)Lin, Liu, Bai, Jin, Li, Zheng, and
  Cao]{lin2026rethinking}
Hongqiang Lin, Chao Liu, Xiaofan Bai, Xuan Jin, Yuhong Li, Nenggan Zheng, and
  Xipeng Cao.
\newblock Rethinking self-evolution: A constrained exploration-exploitation
  process for mitigating skill overfitting.
\newblock \emph{arXiv preprint arXiv:2607.26643}, 2026.

\bibitem[Madaan et~al.(2023)Madaan, Tandon, Gupta, Hallinan, Gao, Wiegreffe,
  Alon, Dziri, Prabhumoye, Yang, Gupta, Majumder, Hermann, Welleck,
  Yazdanbakhsh, and Clark]{madaan2023selfrefine}
Aman Madaan, Niket Tandon, Prakhar Gupta, Skyler Hallinan, Luyu Gao, Sarah
  Wiegreffe, Uri Alon, Nouha Dziri, Shrimai Prabhumoye, Yiming Yang, Shashank
  Gupta, Bodhisattwa~Prasad Majumder, Katherine Hermann, Sean Welleck, Amir
  Yazdanbakhsh, and Peter Clark.
\newblock Self-refine: Iterative refinement with self-feedback.
\newblock In \emph{Advances in Neural Information Processing Systems}, 2023.
\newblock arXiv:2303.17651.

\bibitem[Shen et~al.(2026)Shen, Li, and Zhang]{skilloptlite2026}
Yifei Shen, Bo~Li, and Xinjie Zhang.
\newblock Skillopt-lite: Better and faster agent self-evolution with one line
  of vibe.
\newblock \emph{arXiv preprint arXiv:2607.03451}, 2026.

\bibitem[Shi et~al.(2026)Shi, Zytek, Razavi, Narasimhan, and
  Barres]{shi2026tauknowledge}
Quan Shi, Alexandra Zytek, Pedram Razavi, Karthik Narasimhan, and Victor
  Barres.
\newblock $\tau$-knowledge: Evaluating conversational agents over unstructured
  knowledge.
\newblock \emph{arXiv preprint arXiv:2603.04370}, 2026.

\bibitem[Shinn et~al.(2023)Shinn, Cassano, Berman, Gopinath, Narasimhan, and
  Yao]{shinn2023reflexion}
Noah Shinn, Federico Cassano, Edward Berman, Ashwin Gopinath, Karthik
  Narasimhan, and Shunyu Yao.
\newblock Reflexion: Language agents with verbal reinforcement learning.
\newblock In \emph{Advances in Neural Information Processing Systems}, 2023.
\newblock arXiv:2303.11366.

\bibitem[{Sierra Research}(2026{\natexlab{a}})]{tau2benchrepo}
{Sierra Research}.
\newblock {$\tau$-bench} repository (tau2-bench), commit 363133a.
\newblock \url{https://github.com/sierra-research/tau2-bench},
  2026{\natexlab{a}}.
\newblock Commit 363133a, 29 July 2026.

\bibitem[{Sierra Research}(2026{\natexlab{b}})]{taubenchleaderboard}
{Sierra Research}.
\newblock {$\tau$-bench Leaderboard: $\tau^3$-Banking}.
\newblock \url{https://taubench.com/leaderboard/?benchmark=knowledge},
  2026{\natexlab{b}}.
\newblock Accessed 12 October 2026.

\bibitem[{SpreadsheetBench Team}(2026{\natexlab{a}})]{ssb2dataset}
{SpreadsheetBench Team}.
\newblock {SpreadsheetBench-v2} (dataset).
\newblock \url{https://huggingface.co/datasets/KAKA22/SpreadsheetBench-v2},
  2026{\natexlab{a}}.
\newblock Accessed 13 October 2026.

\bibitem[{SpreadsheetBench Team}(2026{\natexlab{b}})]{ssb2leaderboard}
{SpreadsheetBench Team}.
\newblock {SpreadsheetBench Leaderboard (V2 -- Full)}.
\newblock \url{https://spreadsheetbench.github.io/}, 2026{\natexlab{b}}.
\newblock Accessed 12 October 2026.

\bibitem[Waldin(2026)]{hone2026}
T.~Waldin.
\newblock Hone: Agentic harness optimization on the gepa api.
\newblock \url{https://github.com/twaldin/hone}, 2026.

\bibitem[Wang et~al.(2023)Wang, Xie, Jiang, Mandlekar, Xiao, Zhu, Fan, and
  Anandkumar]{voyager2023}
Guanzhi Wang, Yuqi Xie, Yunfan Jiang, Ajay Mandlekar, Chaowei Xiao, Yuke Zhu,
  Linxi Fan, and Anima Anandkumar.
\newblock Voyager: An open-ended embodied agent with large language models.
\newblock \emph{arXiv preprint arXiv:2305.16291}, 2023.

\bibitem[Wang et~al.(2024)Wang, Mao, Fried, and Neubig]{awm2024}
Zora~Zhiruo Wang, Jiayuan Mao, Daniel Fried, and Graham Neubig.
\newblock Agent workflow memory.
\newblock \emph{arXiv preprint arXiv:2409.07429}, 2024.

\bibitem[Yang et~al.(2026)Yang, Gong, Huang, Yang, Zhou, Huang, Li, Gao, Dai,
  Liu, Qiu, Yang, Chen, Yang, and Luo]{skillopt2026}
Yifan Yang, Ziyang Gong, Weiquan Huang, Qihao Yang, Ziwei Zhou, Zisu Huang, Yan
  Li, Xuemei Gao, Qi~Dai, Bei Liu, Kai Qiu, Yuqing Yang, Dongdong Chen, Xue
  Yang, and Chong Luo.
\newblock Skillopt: Executive strategy for self-evolving agent skills.
\newblock \emph{arXiv preprint arXiv:2605.23904}, 2026.

\bibitem[Zhang et~al.(2026)]{coevoskills2026}
Hanrong Zhang et~al.
\newblock Coevoskills: Self-evolving agent skills via co-evolutionary
  verification.
\newblock \emph{arXiv preprint arXiv:2604.01687}, 2026.

\bibitem[Zhang et~al.(2025)Zhang, Hu, Upasani, Ma, Hong, Kamanuru, Rainton, Wu,
  Ji, Li, Thakker, Zou, and Olukotun]{ace2025}
Qizheng Zhang, Changran Hu, Shubhangi Upasani, Boyuan Ma, Fenglu Hong,
  Vamsidhar Kamanuru, Jay Rainton, Chen Wu, Mengmeng Ji, Hanchen Li, Urmish
  Thakker, James Zou, and Kunle Olukotun.
\newblock Agentic context engineering: Evolving contexts for self-improving
  language models.
\newblock \emph{arXiv preprint arXiv:2510.04618}, 2025.

\bibitem[Zhao et~al.(2023)Zhao, Huang, Xu, Lin, Liu, and Huang]{expel2023}
Andrew Zhao, Daniel Huang, Quentin Xu, Matthieu Lin, Yong-Jin Liu, and Gao
  Huang.
\newblock Expel: {LLM} agents are experiential learners.
\newblock \emph{arXiv preprint arXiv:2308.10144}, 2023.

\bibitem[Zheng et~al.(2025)Zheng, Fatemi, Jin, Wang, Gandhi, Song, Gu,
  Srinivasa, Liu, Neubig, and Su]{skillweaver2025}
Boyuan Zheng, Michael~Y. Fatemi, Xiaolong Jin, Zora~Zhiruo Wang, Apurva Gandhi,
  Yueqi Song, Yu~Gu, Jayanth Srinivasa, Gaowen Liu, Graham Neubig, and Yu~Su.
\newblock Skillweaver: Web agents can self-improve by discovering and honing
  skills.
\newblock \emph{arXiv preprint arXiv:2504.07079}, 2025.

\bibitem[Zhu et~al.(2026)Zhu, Zhang, Ma, Zhang, Schoepf, Woloch, Wang, Yang,
  Jacob, Nagisetty, Chundru, Lin, Mateega, and Zhang]{zhu2026spreadsheetbench2}
Jian Zhu, Yuzheng Zhang, Zeyao Ma, Bohan Zhang, Armin Schoepf, Daniel Woloch,
  Peter~Yiliu Wang, Guangyu~Robert Yang, Samuel Jacob, Siddharth Nagisetty,
  Abhiram Chundru, Jean Lin, Spencer Mateega, and Jing Zhang.
\newblock {SpreadsheetBench} 2: Evaluating agents on end-to-end business
  spreadsheet workflows.
\newblock \emph{arXiv preprint arXiv:2606.29955}, 2026.

\end{thebibliography}

\appendix
\renewcommand\thesubsection{\arabic{subsection}}
\section*{Appendix}
\phantomsection\addcontentsline{toc}{section}{Appendix}\label{sec:appendix-current}
\setcounter{subsection}{0}

\subsection{The SpreadsheetBench~v2 split and grader}
\label{sec:appendix-ssb-split-grader}

\paragraph{SpreadsheetBench~v2: the split.}
Per category (Debugging / Financial\_Model / Template), the $57/57/104$ split is $19/20/18$ training, $20/17/20$ validation and $28/36/40$ held-out tasks. Tasks that share a reference workbook, a Financial\_Model formula family or a Template folder are in one set.

\paragraph{SpreadsheetBench~v2: the grader.}
The grader is the benchmark's evaluation script, which recalculates nothing, with a formula recalculation added (\S\ref{sec:discussion-grader}): it opens the solver's saved workbook in LibreOffice with every formula recomputed, then compares it with the reference workbook; a match scores $1$, anything else $0$. Where LibreOffice cannot compute a formula that the solver left unchanged, the grader uses the value stored in the input workbook. A failed recalculation is a grader failure, not a score.

\subsection{The SpreadsheetBench~v2 baselines' setup}
\label{sec:appendix-ssb-baselines}
For both baselines the solver is a Sonnet~5.5 Claude Code agent, and the held-out test runs $3$ attempts per task. \skillopt{}'s optimizer makes single calls to Claude Sonnet~5.5 through its \texttt{claude\_chat} backend (one \texttt{claude -p} call per proposal or merge); SkillBoost's optimizer is a Sonnet~5.5 Claude Code agent session in each round.

We follow the released code of both baselines except for the following:
\begin{itemize}
\item \textbf{Both:} our grader, which recalculates every formula, replaces each baseline's own SpreadsheetBench scorer.
\item \textbf{SkillBoost:} a $60$-minute limit on its optimizer session, beside its own $80$-turn limit.
\item \textbf{SkillBoost:} the passed tasks each candidate also runs on (up to $10$) are drawn at random each round; its code takes them as an input.
\item \textbf{SkillBoost:} its limit on how much a candidate may grow is lifted while the current skill is empty.
\item \textbf{SkillBoost:} the selection thresholds are set to more training tasks passed and fewer than $5\%$ of them (at most $2$) newly failed; with its default, a candidate that gains nothing would pass.
\end{itemize}

\Needspace*{417pt}%
\subsection{The benchmark adaptor's guidance (appended to Claude Code's system prompt)}
\label{sec:appendix-ba-guidance}
The BA (Claude Sonnet~5.5) ran once for each benchmark, SpreadsheetBench~v2 and \tauthree{}, with this text unchanged.
\begin{lstlisting}[basicstyle=\ttfamily\scriptsize]
You are the benchmark adapter of a skill-evolution run. You turn the user instruction into one file that describes the benchmark to the loop (the program that drives the run) and one directory the evolver will use.

## What a skill-evolution run does

The run evolves a skill set for a solver agent. A skill is a folder the solver can load while it works: instructions and scripts it can run. Each epoch, the solver attempts the training tasks with the current skill set; then an evolver agent reads the traces of these attempts and the grader's scores and writes the next skill set. The tasks the run uses should be divided into three sets:

- **train**: the tasks the skills are learned from. The solver runs them every epoch, and the evolver reads those attempts and everything you keep for it (below).
- **validation**: the tasks that measure each epoch's skill set. The solver runs them every epoch; the evolver sees nothing of them.
- **test**: the tasks held out until the end, when the solver is measured on them once with the final skill set and once without skills.

If the user instruction gives no train/validation/test split, create one yourself. Do not leak any validation or test answer to anything the evolver has access to.

## Your job

Write `benchmark_plugin.py` into `{setup_dir}/` and fill `{reference_dir}/` (use your working directory for anything else):

- `benchmark_plugin.py`: a Python module that describes the benchmark to the loop. It defines one function: `list_tasks(root)`, called with the data root as `root`, returns one `Task` (defined in `plugin_api.py`) for every task the run uses, each with its set: train, validation or test. It reads only files under `root`, building every path as `root/...`; anything else it needs, such as the split, goes in its own source. It imports only the Python standard library and `plugin_api`. It never runs the grader, starts a process or uses the network. The loop grades every rollout with the user's grader, `benchmark_grader.py` in `{setup_dir}/`; fill each `Task` with what that grader reads.
- `{reference_dir}/`: everything the user handed over that the evolver may use. Copy into it every artifact you were given (the data, documentation, anything else) except two kinds: anything that relates to validation or test tasks, and the grader itself, meaning its code and anything that describes how it grades. The evolver gets only scores from the loop's grader and never needs to know how it works. Keep the rest whole and in its original layout. The evolver sees this directory as its only view of the benchmark beyond the solver's attempts and their scores; it may read and run anything in it. **Nothing** in it, whether you copied it or wrote it, may contain either of those two kinds.

## What you can read

`plugin_api.py`, a skeleton plug-in and the user's grader `benchmark_grader.py` are in `{setup_dir}/`. You can also read the data root and every other file the user instruction names, at the paths it gives, the grader's code included. Test your plug-in on the data root before you finish.
\end{lstlisting}

\Needspace*{8\baselineskip}
\subsection{The user instruction for SpreadsheetBench~v2 (the benchmark adaptor's message)}
\label{sec:appendix-instruction-ssb}
\begin{lstlisting}[basicstyle=\ttfamily\scriptsize]
### 1. Benchmark, data and split

SpreadsheetBench v2, the three non-Visualization categories: Debugging, Financial_Model and Template (297 tasks: 100, 100 and 97). The data root also holds a fourth folder, Visualization, which is not part of this run. A task gives the solver a workbook and an instruction; the solver edits the workbook with Python and saves it.

The data root is `/data/spreadsheetbench-v2`, with one `dataset.json` per category; each record has `id`, `instruction`, `spreadsheet_path`, `golden_response_path` and `answer_position`; the two paths are relative to that category's folder. Several records share one golden_response_path (a Debugging project has ten tasks on one reference workbook, a Financial_Model project five); tasks that share a reference workbook share an answer, and no such group spans two of train, val and test. A task id is the category plus the record id (`<Category>/<NN_MM>`).

Use the given split: `/data/split.json`, 57 train / 57 val / 104 test.

### 2. What the solver gets and must produce

Per task, the solver gets the instruction and the input workbook in its working directory. It must save the modified workbook to the output path it is given.

### 3. Grader

The grader is `benchmark_grader.py`, given next to `plugin_api.py`. Its `grade(task, snapshot)` scores one rollout from 0 to 1; the snapshot is the solver's working directory when it exited. It reads the task's `inputs[0]` (the input workbook), `answer_files[0]` (the reference workbook) and `grading["answer_position"]`. The loop runs it; you do not need to.

### 4. Environment

The solver and the grader run in the image built from `scripts/daytona/Dockerfile.ssb_leg` in the same repository. The solver's Python is 3.11 with openpyxl 3.1.3, pandas 2.2.0 and numpy 1.26.4.

### 5. Run size

3 epochs · 3 samples per task.
\end{lstlisting}

\Needspace*{8\baselineskip}
\subsection{The user instruction for \texorpdfstring{\tauthree{}}{tau3-bench} (the benchmark adaptor's message)}
\label{sec:appendix-instruction-tau3}
\begin{lstlisting}[basicstyle=\ttfamily\scriptsize,escapeinside={(*}{*)}]
### 1. Benchmark, data and split

(*$\tau^3$*)-bench, the banking domain only (`banking_knowledge`). A task is a conversation between a bank's customer service agent (the solver) and a simulated customer who has a goal; the agent handles it according to the bank's policy, using the bank's tools and its knowledge base.

The data root is `/data/tau3_data_363133a`. `tau2/domains/banking_knowledge/` holds `db.json` (the bank's database), `documents/` (the knowledge base, 698 documents), `prompts/` (the agent's policy texts) and `tasks/` with one file per task, `task_NNN.json` (the customer's scenario, the database's starting state and how the task is judged). `tau2/user_simulator/` holds the customer simulator's guidelines. `.embeddings_cache/` at the data root holds the knowledge base's precomputed embeddings (OpenAI `text-embedding-3-large`, 698 documents), used by the dense search. A task id is `banking_knowledge/` plus the task file's name without `.json` (e.g. `banking_knowledge/task_004`).

Use the given split: `/data/split.json`, 19 train / 20 validation / 31 test.

### 2. What the solver gets and must produce

Per task, the solver is the bank's agent in one conversation. It gets the bank's policy for the `alltools` knowledge-base setup (search with `KB_search_bm25` (keyword) and `KB_search_dense` (embedding), and a read-only `shell` over the knowledge base) and the customer's messages, and it acts through the bank's tools. When the conversation ends, the conversation and the bank's final database are saved under `tau2/` in the solver's working directory.

### 3. Grader

The grader is `benchmark_grader.py`, given next to `plugin_api.py`. Its `grade(task, snapshot)` scores one rollout from 0 to 1; the snapshot is the solver's working directory when the conversation ended. It reads the task's `answer_files[0]` (the task file), `grading["domain"]` (`banking_knowledge`) and `grading["retrieval_variant"]` (`alltools`); a task has no input files.

### 4. Run size

3 epochs · 3 samples per task.
\end{lstlisting}

\Needspace*{8\baselineskip}
\subsection{The evolver's appended system prompt}
\label{sec:appendix-evolver-system}
Appended once to Claude Code's own system prompt and kept for the whole run; the loop fills in one placeholder, the number of attempts per training task. Every \evoskillv{} run in this report used this text.
\begin{lstlisting}[basicstyle=\ttfamily\scriptsize]
You are the skill evolver in a multi-epoch run. Each epoch, a separate solver agent runs every training task 3 times (each run is a rollout); you read what happened and write the skill set it carries next. The goal is a skill set that raises the solver's score on tasks it has never seen. The training tasks are a sample of a larger benchmark, so a skill that describes them specifically will not transfer.

Epoch 0 ran with no skills; each later epoch K runs with `skills_vK`, the set you write at your turn K. This session lasts the whole run: each epoch sends you one message, and your reply to it is your turn.

- `/evo/traces/epoch_K/<task id>_s<i>.jsonl`: the solver's session log of sample i; files the log refers to (such as large tool outputs) are in `<task id>_s<i>/`.
- `/evo/scores/epoch_K.json`: each rollout's score from the grader, 0 to 1, or `grader_failed` when the grader itself failed.
- `/evo/solver_workspaces/epoch_K/<task id>_s<i>/`: the solver's working directory as it was when it exited.
- `/evo/skills/epoch_K/`: `skills_vK`, the set epoch K's solver carried.
- `/evo/staging/`: the set you are writing. Each turn starts with the set you left here; edit it in place, and whatever is here when your turn ends is the next set. If you change nothing, the solver carries the same set again. Each skill is `<skill-name>/SKILL.md`; write only `name` and `description` in its YAML frontmatter. Files beside it ship with the skill, read-only and not executable (the solver runs scripts through their interpreter).
- `/evo/evolver_workspace/`: your own directory, kept across epochs. Your context may be auto-compacted when it grows too big, so keep here anything you want to preserve.
- `/evo/reference/`: the benchmark's data and documents, without anything about validation or test tasks or the grader.

Paths in the traces refer to the solver's machine, not this one; its working directory and skills are the snapshots above. You run in the solver's image, with no network beyond the model API.
\end{lstlisting}

\Needspace*{8\baselineskip}
\subsection{The evolver's message at each epoch}
\label{sec:appendix-evolver-message}
The loop fills in the current time, the epoch's deadline (two hours later) and the epoch number. As sent at epoch~$1$ of the \tauthree{} run with training answers:
\begin{lstlisting}[basicstyle=\ttfamily\scriptsize]
It is 2026-10-07 21:22 UTC; you have until 2026-10-07 23:22 UTC for this epoch, after which `/evo/staging/` is taken as it stands. You are at epoch 1. Please analyze the solver rollouts, diagnose the pitfalls that may be fixed by a dedicated skill, and compose `skills_v1` under `/evo/staging/`.
\end{lstlisting}

\Needspace*{8\baselineskip}
\subsection{The SpreadsheetBench~v2 evolver's skills (fine-grained ground truth): their descriptions}
\label{sec:appendix-current-skills}\label{sec:appendix-ssb-skills}
The four epoch-3 skills of the \evoskillv{} run with fine-grained ground truth (\S\ref{sec:results-ssb}, Figure~\ref{fig:heldout-ssb}). The Sonnet~5.5 evolver wrote all four at its first turn, kept them to the last and revised each one at every turn; at epoch~$3$ their \texttt{SKILL.md} files hold $71$\,KB of text. \texttt{debug-injected-errors} ships three scripts that only report (suspicious formulas; each row's formulas grouped by pattern; the links between sheets), and \texttt{financial-model-fill} one that finds the blanked cells and one that lists the input cells no formula uses. \texttt{bond-deferred-tax-templates} (Template folders $01$ and $05$) and \texttt{re-direct-cap-model} (folder $16$) are written for the only three Template folders in the training set, from which no validation or held-out task comes. Each skill's name and description, as the evolver wrote them (the file names they cite belong to training tasks):
\begin{lstlisting}[basicstyle=\ttfamily\scriptsize]
name: debug-injected-errors
description: Fix deliberately injected errors in a finished multi-sheet financial model (LBO, Operating Model + DCF, Revenue Build, Comps + WACC and similar). Use when the input file name or task text names an error category - Double Counting, Cross Sheet references, Index Match, AVERAGE, Embedded Hardcodes, Unit Mismatch, Sign Conventions, Relative vs Absolute references, Inconsistent Color Coding, or "Errors" (#REF!/deleted rows). Gives a precision-first minimal-edit workflow, the exact mutation signatures per category, original-model quirks that must NOT be fixed, and three scanner scripts.

name: financial-model-fill
description: Fill blanked-out cells in multi-sheet corporate financial model workbooks (assumptions, revenue drivers, income statement, balance sheet, cash flow, WC/debt/PPE schedules, DCF/WACC, valuation multiples, ratios, sensitivity tables) with live Excel formulas via openpyxl so that the LibreOffice-recalculated values match a hidden golden model exactly. Use when a task says "complete/populate/fill the cells in sheet X" for a project model. LOAD FIRST for every 'Complete the financial model based on the provided assumptions' task on a corporate project workbook (file names like 01_0X_Education, 06_DigiMark, 14_Refinery, 16_Ramjet, 20_Fintech); do not start editing before reading it.

name: bond-deferred-tax-templates
description: Conventions for completing "Bond accounting" (effective-interest, OID, zero-coupon, premium/discount, early extinguishment, GAAP vs tax balance sheet) and "Deferred tax" (book vs taxable income, DTA/DTL, cash-flow statement) template workbooks. Use when a sheet has labels like "Interest expense (amortization)", "Loan balance - BOP", "Deferred tax asset - BOP", "Tax B/S impact of borrowing", "Implied IRR", or "Balance check" / "Unearned revenue". LOAD FIRST for any Template task on bond/loan/deferred-tax/GAAP-vs-tax workbooks (file names 01_0X_input.xlsx, 05_0X).

name: re-direct-cap-model
description: LOAD FIRST for any workbook with sheets named DirectCap, RentRoll, NOI_Analysis, Valuation, Assumptions/SharedData or labels like Total Units, Average Unit Size (SF), Per SF/Month, Per Unit, Effective Gross Revenue, NET OPERATING INCOME, Cap Rate, Property Valuation / Indicated Value (file names 16_0X_input.xlsx, 'Complete the financial model based on the provided assumptions. Ensure that all formulas, references, and calculations are correct.'). Multifamily real-estate direct-capitalization models - rent roll totals, pro forma revenue, vacancy, opex, NOI, cap rate, indicated value, per-SF/per-unit/per-month columns (including the NOI row of a per-SF column). Template tasks are graded cell by cell; unlike the corporate-model skill this one lists the exact conventions.
\end{lstlisting}

\subsection{The \texorpdfstring{\tauthree{}}{tau3-bench} banking evolver's skills (fine-grained ground truth): their descriptions}
\label{sec:appendix-tau3-skills}
The two epoch-3 skills of the \evoskillv{} run with fine-grained ground truth and the Sonnet~5.5 evolver (\S\ref{sec:results-tau3}, Figure~\ref{fig:heldout-tau3}). Their \texttt{SKILL.md} files hold $11$\,KB of text and ship no scripts.
\begin{lstlisting}[basicstyle=\ttfamily\scriptsize]
name: rho-bank-kb-workflow
description: Workflow for Rho-Bank customer-service tasks that are scored on exact tool actions (verification, discoverable tools, closures, replacements, applications, transfers). Use at the start of any Rho-Bank / knowledge-base banking support conversation.

name: rho-bank-pitfalls
description: Concrete failure patterns observed in Rho-Bank KB tasks (credit card closure/retention, debit card replacement, ATM limits, savings APY, cash-back disputes, card promos). Read when handling these scenarios.
\end{lstlisting}

\end{document}